\documentclass[10pt,twocolumn,letterpaper]{article}

\usepackage{cvpr}
\usepackage{times}
\usepackage{epsfig}
\usepackage{graphicx}
\usepackage{amsmath}
\usepackage{amssymb}
\usepackage{subfig}
\usepackage{algorithm}
\usepackage{algpseudocode}
\usepackage{algorithmicx}
\usepackage{arydshln}
\usepackage{multirow}
\usepackage{xcolor}
\usepackage{authblk}

\usepackage[breaklinks=true,bookmarks=false]{hyperref}

\cvprfinalcopy 

\def\cvprPaperID{****} 

\begin{document}

\title{Heavy Rain Image Restoration: Integrating Physics Model and Conditional Adversarial Learning\thanks{This work is supported by the DIRP Grant R-263-000-C46-232.  R.T. Tan's  research is supported in part by Yale-NUS College Start-Up Grant.}}

\author[1]{Ruoteng Li}
\author[1]{Loong-Fah Cheong}
\author[1,2]{Robby T. Tan}
\affil[1]{National University of Singapore} \affil[2]{Yale-NUS College}

\maketitle

\begin{abstract}
Most deraining works focus on rain streaks removal but they cannot deal adequately with heavy rain images.  In heavy rain, streaks are strongly visible, dense rain accumulation or rain veiling effect significantly washes out the image, further scenes are relatively more blurry, etc. In this paper, we propose a novel method to address these problems. We put forth a 2-stage network: a physics-based backbone followed by a depth-guided GAN refinement. The first stage estimates the rain streaks, the transmission, and the atmospheric light governed by the underlying physics. To tease out these components more reliably, a guided filtering framework is used to decompose the image into its low- and high-frequency components. This filtering is guided by a rain-free residue image --- its content is used to set the passbands for the two channels in a spatially-variant manner so that the background details do not get mixed up with the rain-streaks. For the second stage, the refinement stage, we put forth a depth-guided GAN to recover the background details failed to be retrieved by the first stage, as well as correcting artefacts introduced by that stage. We have evaluated our method against the state of the art methods. Extensive experiments show that our method outperforms them on real rain image data, recovering visually clean images with good details.
\end{abstract}

\section{Introduction}
\begin{figure}
  \centering
  \subfloat[Input Image]{\includegraphics[width=0.49\linewidth]{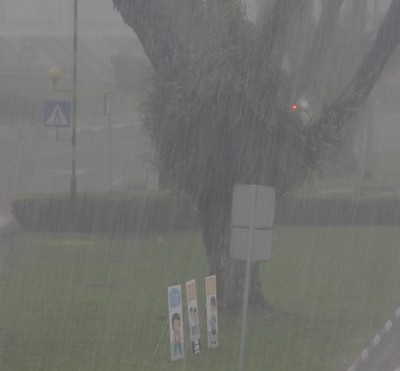}}
  \subfloat[Our Result]{\includegraphics[width=0.49\linewidth]{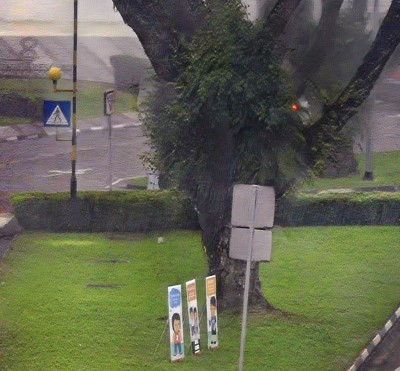}} \\ \vspace{-5pt}
  \subfloat[Non-Local\cite{NonLocalImageDehazing}+RESCAN\cite{Li_2018_ECCV}]{\includegraphics[width=0.49\linewidth]{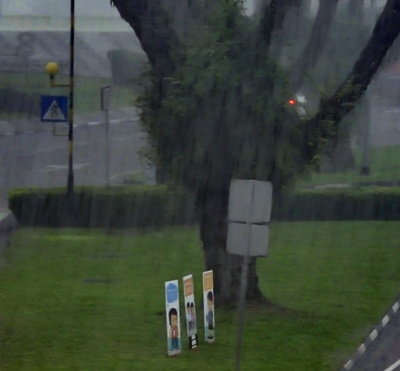}}
  \subfloat[Non-Local\cite{NonLocalImageDehazing}+DIDMDN\cite{Zhang_2018_CVPR}]{\includegraphics[width=0.49\linewidth]{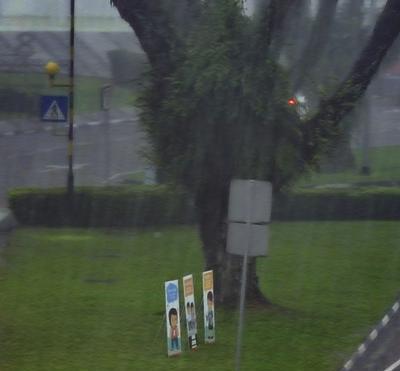}} \\
  \caption{A comparison of our algorithm with combined state of the art dehazing/defogging \cite{NonLocalImageDehazing} and deraining \cite{Li_2018_ECCV}\cite{Zhang_2018_CVPR}. (Zoom-in to view details.)}\label{fig:sample}
\end{figure}

As one of the commonest dynamic weather phenomena, rain causes  significant detrimental impacts on many computer vision algorithms \cite{Richter_2017}. A series of rain removal methods have been proposed to address the problem (e.g. \cite{Kang12Rain, Jiang_2017_CVPR,Zhang:2006:Derain,Fu_2017_CVPR,YangTFLGY16,Li_2016_CVPR,Wei_2017_ICCV,Zhu_2017_ICCV,Liu_2018_CVPR,Chen_2018_CVPR,Ren_2017_CVPR,Li_2018_ECCV}). Principally, these methods rely on the following model:
\begin{equation}\label{eq:simple-rain-model}
  \mathbf{I} = \mathbf{J} + \sum_{i}^{n} \mathbf{S}_i,
\end{equation}
where $\mathbf{I}$ is the observed input image. $\mathbf{J}$ is the background scene free from rain. $\mathbf{S}_i$ is the rain layer, with $n$ as the total number of rain-streak layers.

While the model in Eq.~(\ref{eq:simple-rain-model}) is widely used, it crudely represents the reality. In real rain, particularly in relatively heavy rain, aside from the rain streaks, there is also a strong veiling effect, which is the result of rain-streak accumulation in the line of sight. This important rain veiling effect (also known as rain accumulation) is ignored in the model. Hence, most of the existing methods do not perform adequately when dense rain accumulation is present (shown in Fig.~\ref{fig:sample}). As one can observe in the figure, a state of the art method of rain-streak removal \cite{Li_2018_ECCV} combined with a state of the art dehazing/defogging method \cite{NonLocalImageDehazing} still retains some rain streaks and veiling effect in the output. Note, zooming in the image will reveal the streaks and veiling effect.

The density of rain, both rain streaks and accumulation, is a spectrum. Thus, there is no clear dividing line between what light and heavy rain are. In this paper, we associate heavy rain to the severity of its visual degradation, namely when the rain streaks are strongly visible, the veiling effect significantly washes out the image, the distant background scenes are slightly blurry (due to multiflux scattering), and the physical presence of the rain streaks and rain accumulation is entangled with each other. The purpose of using the term ``heavy rain'' is to differentiate our method from other methods that do not address the mentioned problems.

To achieve our goal of restoring an image degraded by heavy rain, we need to address a few problems related to it. First, we can no longer utilize the widely used model (Eq.~(\ref{eq:simple-rain-model})), since it does not accommodate rain accumulation. We need a model that can represent both rain streaks and rain accumulation, like the one introduced by \cite{YangTFLGY16}:
\begin{equation}
\label{eq:full_rain_model}
\mathbf{I} = \mathbf{T} \odot (\mathbf{J} + \sum_{i}^{n} \mathbf{S_i}) + (\mathbf{1} - \mathbf{T}) \odot \mathbf{A},
\end{equation}
where $\mathbf{T}$ is the transmission map introduced by the scattering process of the tiny water particles, $\mathbf{A}$ is the global atmospheric light of the scene. $\mathbf{1}$ is a matrix of ones, and $\odot$ represents element-wise multiplication.

Second, aside from the model, existing methods tend to fail in handling heavy rain because, when dense rain accumulation (dense veiling effect) is present, the appearance of the rain streaks is different from the training data of the existing methods \cite{Fu_2017_CVPR,Zhang_2018_CVPR,YangTFLGY16}. In the real world, rain streaks and rain accumulation can entangle with  each other, which is intractable to be rendered using simple physics models. Hence, a sequential process (e.g, rain-streak removal followed by rain-accumulation removal) as suggested in \cite{Li_2016_CVPR,YangTFLGY16} cannot solve the problem properly. Moreover, unlike in fog images, estimating the atmospheric light, $\mathbf{A}$, in rain images is more complex, due to the strong presence of rain streaks. Note that, the proper estimation of the atmospheric light is critical, since it affects the restoration outputs significantly.

Third, particularly in heavy rain, the visual information of the background scene can be severely damaged. This is due to both rain streaks and rain accumulation as described in Eq.~(\ref{eq:full_rain_model}). Unfortunately, some of the damages are not represented by the model. One of them is multiflux scattering effect in the form of blurriness of the scenes, particularly the further scenes \cite{narasimhan2003shedding}. In other words, the model cannot fully represent what happens in the real world. This creates performance problems, especially for methods that rely on the model, like most of the methods do.

To address these existing problems resulted by heavy rain, we introduce a novel CNN method to remove rain streaks as well as rain accumulation simultaneously with the following contributions:
\begin{enumerate}

	\item
We introduce an integrated two-stage neural network: a physics-based subnetwork and a model-free refinement subnetwork, to address the gap between physics-based rain model (Eq.~(\ref{eq:full_rain_model})) and real rain. The first stage estimates $\mathbf{S}$, $\mathbf{A}$, $\mathbf{T}$ and produces reconstructed image $\mathbf{J}$ strictly governed by the rain model. The second stage contains a conditional GAN (cGAN) \cite{Mehdi_2016_NIPS}  that is influenced strongly by the outputs of the first stage.

	\item
We propose novel streak-aware decomposition to adaptively separate the image into high-frequency component containing rain streaks and low-frequency component containing rain accumulation. This addresses the problem of entangled appearance of rain streaks and rain accumulation. Also, since we can have a low frequency component, we can utilize it to resolve the problem of estimating the atmospheric light, $\mathbf{A}$.


	\item
We provide a new synthetic data generation pipeline that synthesizes the veiling effect in a manner consistent with the scene depth. For more realism, we also add Gaussian blur on both the transmission map and the background to simulate the effect of scattering in heavy rain scenarios.

\end{enumerate}

Using these ideas, our experimental results show the superiority of our method compared to the state of the art methods qualitatively and quantitatively.

\begin{figure*}
  \centering
  \includegraphics[width=1.0\linewidth]{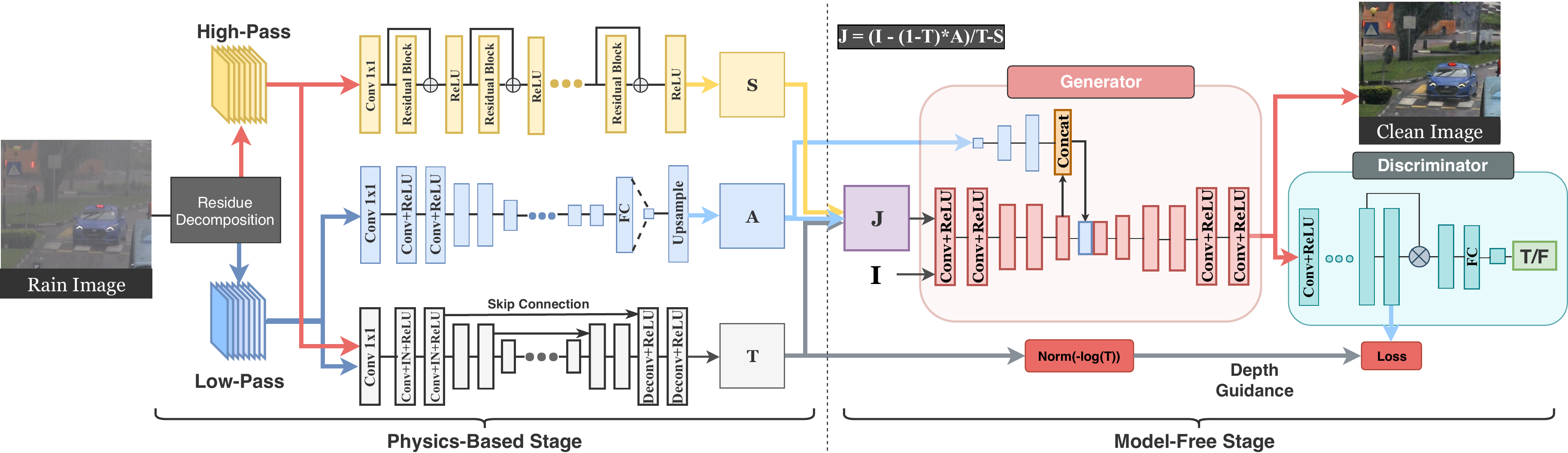}
  \caption{The overall architecture of the proposed network. The details of the residue decomposition module is shown in Fig.~\ref{fig:dm}. The image $\mathbf{J}$ is reconstructed according to Eq.~(\ref{eq:J_reconstruction}) during training. }\label{fig:arch}
\end{figure*}

\section{Related Works}

Most existing deraining methods are not designed for heavy rain scenes, therein lies the main difference with our work. This applies to all the image-based  \cite{Kang12Rain,Luo2017,Jiang_2017_CVPR,Li_2016_CVPR,YangTFLGY16,Fu_2017_CVPR,Zhang_2018_CVPR,Li_2018_ECCV} and video-based works \cite{Zhang:2006:Derain,Garg:2007:VR,barnum2007spatio,Bossu2011,Liu_2018_CVPR,Kim_2015_TIP,Chen_2018_CVPR,Li_2018_CVPR_rain,Varun_SPIE_2012,Chen_TIP_2014,Santh_IJCV_2015,Tripathi_IET_2012,You_TPAMI_2016}. In the following, we focus our review on the image-based works.

Kang et al.'s \cite{Kang12Rain} introduces the very first single image deraining method that decomposes an input image into its low frequency component and a high-frequency component using bilateral filter. The main difference with our decomposition method lies in that its high-frequency layer contains both rain streaks and high-frequency background details---its sparse-coding based method using dictionary cannot differentiate genuine object details from the rain streaks. Li et al.'s \cite{Li_2016_CVPR} decomposes the rain image into a rain-free background layer and a rain streak layer, by utilizing Gaussian Mixture Models (GMMs) as a prior for the background and rain streak layers. This paper also attempts to address rain accumulation using a pre-processing dehazing step \cite{Cai_2016_TIP}. However, the dehazing step enhances clear rain streak further, causing the rain streak's contrast and intensity much higher than that of the training data.  Thus, the subsequent rain streak removal method cannot effectively remove boosted rain streaks. Fu et al. \cite{Fu_2017_CVPR} proposes a deep convolutional network solution that is based on an image decomposition step similar to \cite{Kang12Rain} and the details layer again contain both rain streaks and background details, which hampers the learning of rain streaks. Yang et al.'s \cite{YangTFLGY16} removes the rain accumulation using a dehazing method \cite{Cai_2016_TIP} as an iteration step in his recurrent framework. However in heavy rain scenes, a large number of noise hidden in the atmospheric veils will be boosted by dehazing method, which cannot be handled by Yang et al's rain streak removal module. Without treating the rain accumulation problem in an integral manner like our approach, it can only work well for the veiling effect produced in light rain, but not the heavy rain discussed in this paper. Both \cite{Zhang_2018_CVPR} and \cite{Li_2018_ECCV} are deep learning approaches that attempt to deal with the complex overlaying of rain layers in heavy rain scenes (by being density-aware and by having a recurrent network, respectively) but they do not deal with rain accumulation, and they also fail to remove the rain streaks cleanly in our experiments.

\section{Network Design}

Before describing the proposed 2-stage network, we first discuss the overall input and output of the network, as well as the intermediate output by the first stage. Referring to Fig.~\ref{fig:arch}, the first stage,  the physics-based network, takes in a single rain image as input and extracts the physical parameters of rain, including the rain streak intensity $\mathbf{S}$, atmospheric light $\mathbf{A}$ and transmission $\mathbf{T}$. The output of this first stage is the clean background image $\mathbf{J}$ computed by the following equation (derived from Eq.~(\ref{eq:full_rain_model})):
\begin{equation}\label{eq:J_reconstruction}
\mathbf{J} = {\mathbf{I} -  (\mathbf{1} - \mathbf{T}) \odot \mathbf{A} \over \mathbf{T}} - \sum_{i}^{n}\mathbf{S}_i .
\end{equation}
The cGAN in the second stage refines the estimated  $\mathbf{J}$ to produce the clean background image $\mathbf{C}$ as our final output.

The reason of proposing the 2-stage network is as follows. The physics model (Eq.~(\ref{eq:full_rain_model})) is an approximated representation of real rain scenes, and thus can provide constraints to our network, such as rain-streaks ($\mathbf{S}$), atmospheric light ($\mathbf{A}$), and transmission ($\mathbf{T}$). However, there is a significant disadvantage of using the physics model alone to design the network, since the model is only a crude approximation of the real world. Therefore, using a network that is purely based on the model will not make our method robust, particularly for heavy rain. As mentioned in the introduction, the damages induced by rain streaks and rain accumulation cannot be fully expressed by the model (Eq.~(\ref{eq:full_rain_model})). For this reason, we add another network, the model-free network, which does not assume any model. Hence, unlike the first network, this network has less constraints and adapts more to the data. However, we cannot use this network alone either, since there is no proper guidance to the network in transforming a rain image to its clean image.

\subsection{Stage 1: Physics-based Restoration}
The outline of our physics-based network is as follows. First, it decomposes the input image into high and low frequency components, where from the high frequency component, the network estimates the rain-streaks map, $\mathbf{S}$, and from the low frequency component, it estimates the atmospheric light, $\mathbf{A}$, and the transmission map, $\mathbf{T}$, as shown in Fig.~\ref{fig:arch}. The details of these processes are discussed in these subsequent sections.

\vspace{0.3cm}

\begin{figure}
\centering
  \includegraphics[width=1.0\linewidth]{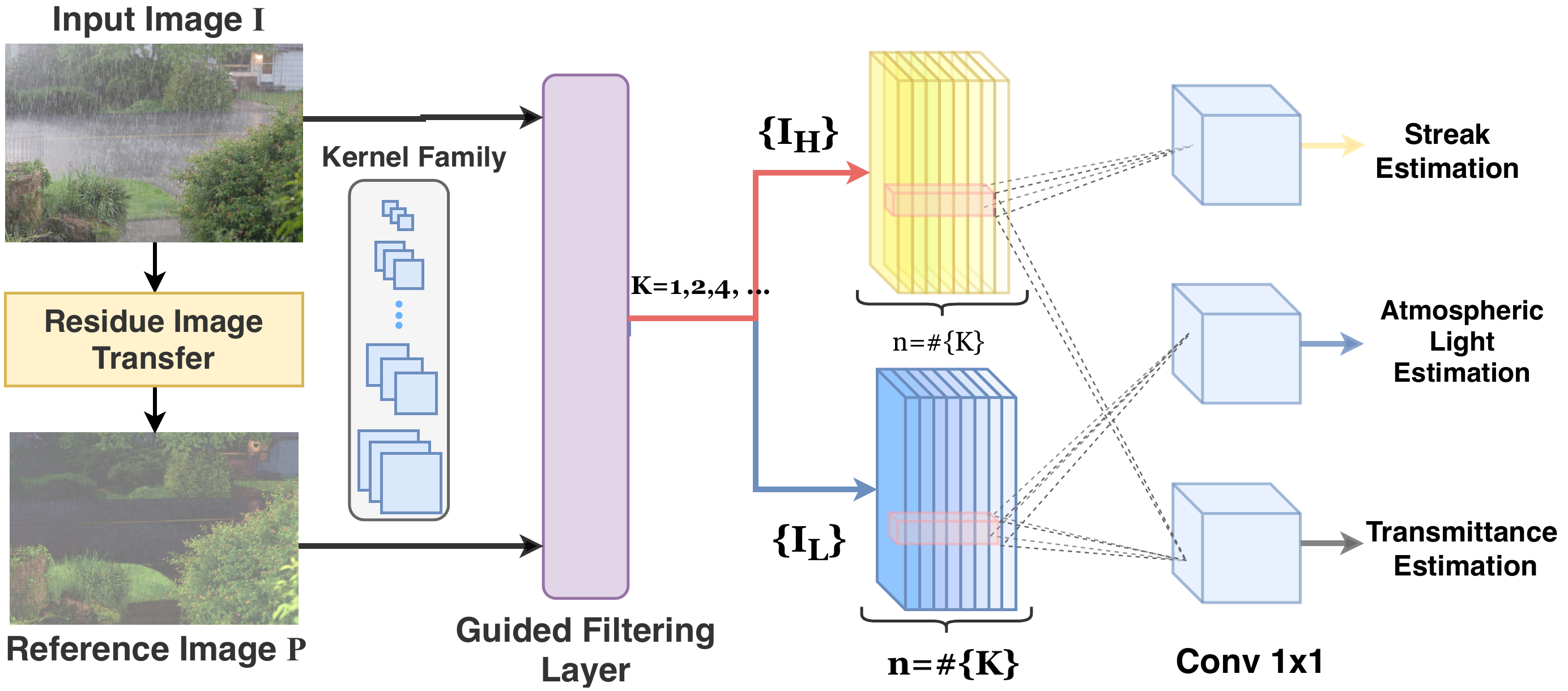}
  \caption{The schematic view of the structure of colored-residue image guided decomposition module.}\label{fig:dm}
  \vspace{-5pt}
\end{figure}

\noindent \textbf{Residue Channel Guided Decomposition }
In rain images, particularly heavy rain, the visual appearances of rain streaks and rain accumulation are entangled in each other. This entanglement causes complexity in estimating the rain parameters:  $\mathbf{S}$,  $\mathbf{A}$, and  $\mathbf{T}$.  Estimating  $\mathbf{A}$ and  $\mathbf{T}$ from the input image directly will be complex due to the strong presence of rain streaks. Similarly, estimating $\mathbf{S}$ from the raw input image is intractable, due to the strong presence of dense rain accumulation. For this reason, we propose a process to decompose the input image into high and low frequency components, to reduce the complexity of the estimations and thus increase the robustness.

Our decomposition is adopted from \cite{Wu_2018_CVPR}, where we create a decomposition CNN layer that is differentiable during training (details shown in Fig.\ref{fig:dm}). Specifically, we first perform image smoothing on the input image $\mathbf{I}$. The smoothed image is considered as the low-frequency component $\mathbf{I}_L$ while the subtraction $\mathbf{I}_H = \mathbf{I} - \mathbf{I}_L$ provides the high-frequency component. In each component, Eq.~(\ref{eq:full_rain_model}) becomes:
\begin{eqnarray}
\nonumber
  \mathbf{I}_{H} &=& (\mathbf{1} - \mathbf{T}_{H}) (\mathbf{J}_{H} + \mathbf{S}_{H}) + \mathbf{T}_{H} \mathbf{A}_{H}, \\
  \mathbf{I}_{L} &=& (\mathbf{1} - \mathbf{T}_{L}) (\mathbf{J}_{L} + \mathbf{S}_{L}) + \mathbf{T}_{L} \mathbf{A}_{L},
\label{eq:HFLF_original}
\end{eqnarray}
where ${(\cdot)_{H}, (\cdot)_{L}}$ represent the high-frequency component and low-frequency component respectively. Assuming the atmospheric light $\mathbf{A}$ is constant throughout the image, we can assume that $\mathbf{A}_{H} = \mathbf{0}$. In addition, we also assume that low-frequency component of rain streak $\mathbf{S}_L$ is negligible, i.e., $\mathbf{S}_L = \mathbf{0}$. In other words, the low frequency of rain streaks mainly manifests itself as a veil (rain accumulation), and is modeled by $\mathbf{A}_{L}$. Hence, Eq.~(\ref{eq:HFLF_original}) reduces to:
\begin{eqnarray}
 \nonumber 
  \mathbf{I}_{H} &=& (\mathbf{1} - \mathbf{T}_{H}) (\mathbf{J}_{H} + \mathbf{S}_{H}), \\
  \mathbf{I}_{L} &=& (\mathbf{1} - \mathbf{T}_{L}) (\mathbf{J}_{L}) + \mathbf{T}_{L} \mathbf{A}_{L}.
  \label{eq:HFLF_reduced}
\end{eqnarray}

\begin{figure}
  \centering
  \subfloat[Rain image]{\includegraphics[width=0.32\linewidth]{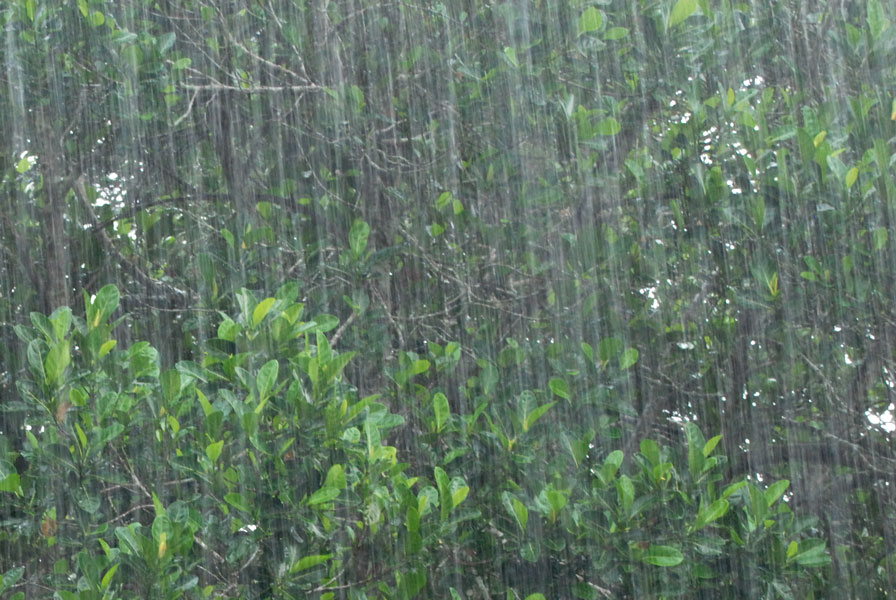}}
  \subfloat[Input-guided $\mathbf{I}_L$]{\includegraphics[width=0.32\linewidth]{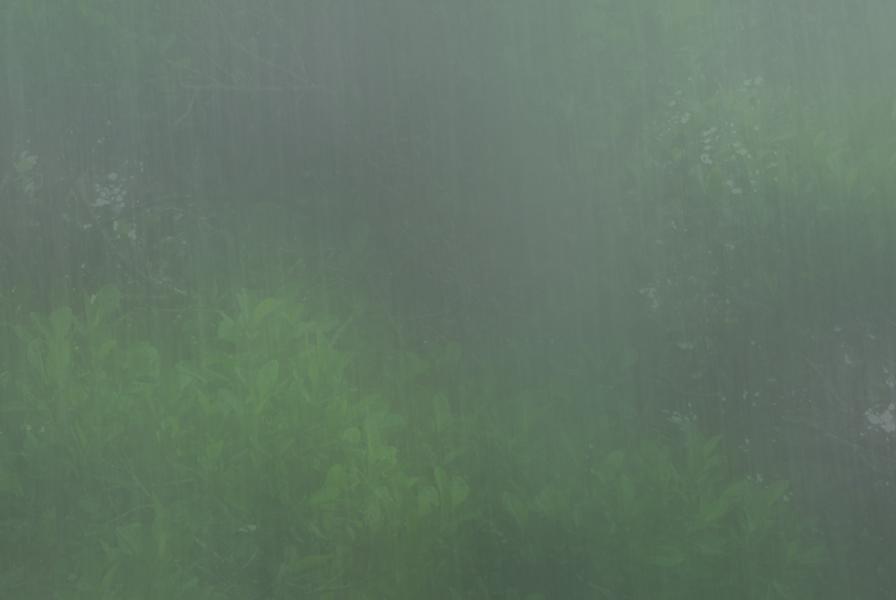}}
  \subfloat[Input-guided $\mathbf{I}_H$]{\includegraphics[width=0.32\linewidth]{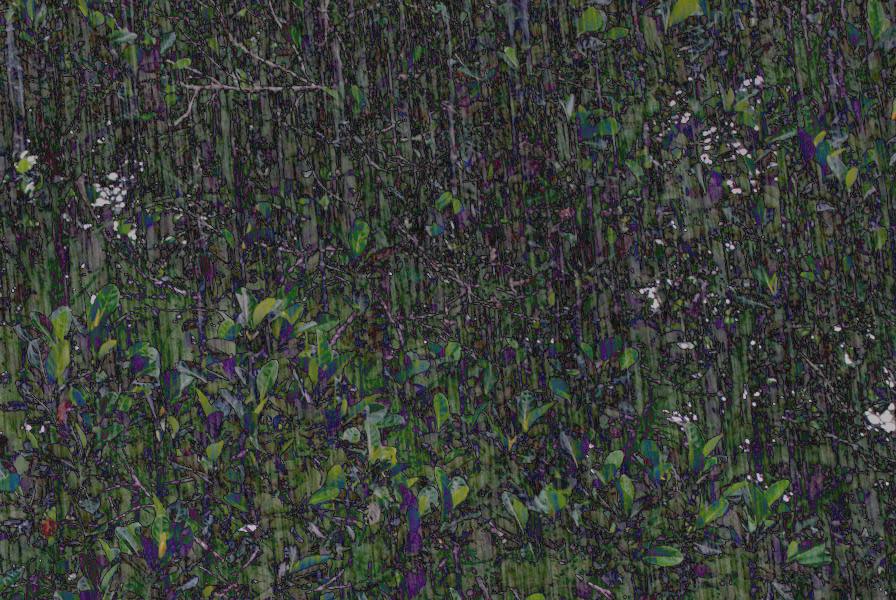}} \\
  \subfloat[Residue channel]{\includegraphics[width=0.32\linewidth]{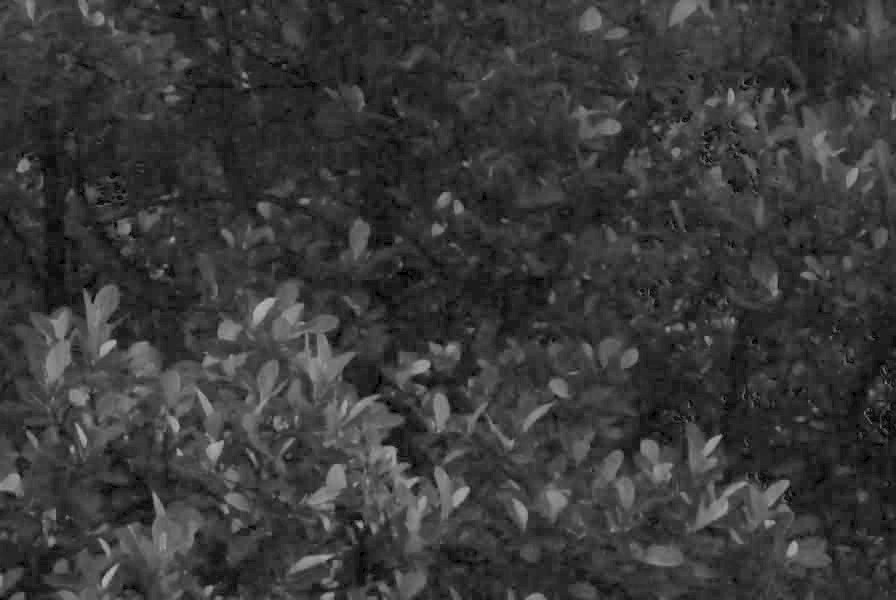}}
  \subfloat[Residue-guided $\mathbf{I}_L$]{\includegraphics[width=0.32\linewidth]{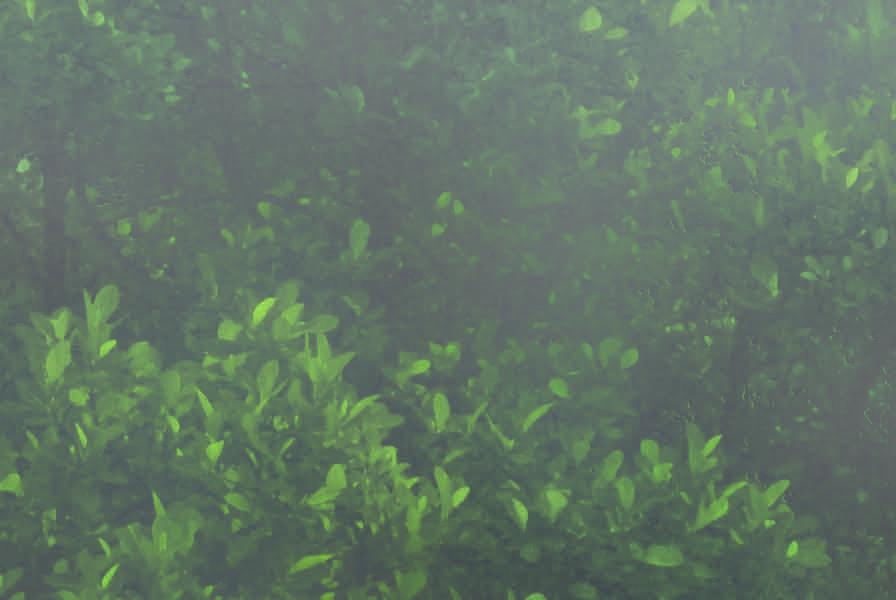}}
  \subfloat[Residue-guided $\mathbf{I}_H$]{\includegraphics[width=0.32\linewidth]{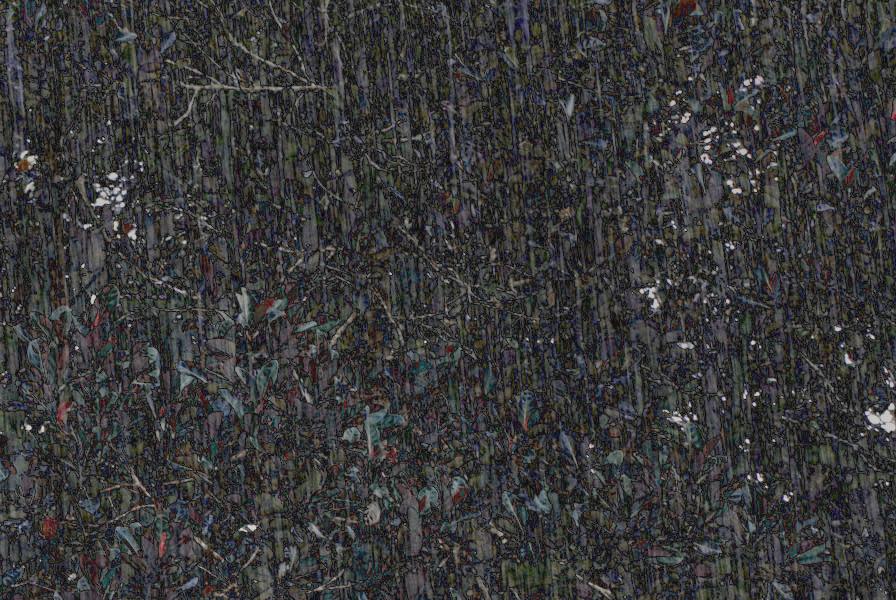}}
  \caption{Input rain image decomposition using (a) input image itself  and (d) its residue channel (kernel size $k = 64 \times 64$) as guidance image. One can observe that more background details are left in the low-frequency channel. }\label{fig:HFLF}
\vspace{-5pt}
\end{figure}

The most important difference in our frequency decomposition lies in the use of the residue image \cite{Li_2018_ECCV_flow} as a reference image to guide the filtering during the aforementioned low-pass smoothing process. This guided filtering allows us to have a spatially variant low-frequency passband that selectively retains the high-frequency background details in the low-frequency channel. As a result, the high-frequency channel contains only rain streaks unmarred by high-frequency background details, which greatly facilitates the learning of rain streaks. The residue image is defined in \cite{Li_2018_ECCV_flow} as follows:
\begin{equation}\label{eq:residue_channel}
  \mathbf{I}_{res}(x) = \max_{c \in {r,g,b}} \mathbf{I}^c(x) - \min_{d \in {r,g,b}} \mathbf{I}^d(x),
\end{equation}
where $\mathbf{I}^c, \mathbf{I}^d$ are the color channels of $\mathbf{I}$. This residue channel is shown to be invariant to rain streaks, i.e., it is free of rain streaks and contains only a transformed version of the background details (see Fig.~\ref{fig:HFLF} (d)). It can thus provide information to guide and vary the passband in the low-frequency smoothing so that the background details are not smoothed away. In practice, we use the colored-residue image \cite{Li_2018_ECCV_flow} as shown in Fig.~\ref{fig:dm}. 

To handle the large variation in the rain streak size present in our rain images, the decomposition uses a set of smoothing kernels ${K}$, with size given by $k = 2^i$, $i= 0,1,...$. In each of the frequency channels, we concatenate these images and send them to a $1 \times 1$ convolutional kernel, which behaves as a channel-wise feature selector.


\vspace{0.3cm}
\noindent \textbf{Learning Rain Streaks}
From the high-frequency component ${\mathbf{I}_H}$, we learn the rain streaks $\mathbf{ S}$ from the ground-truth streaks map using a fully convolutional network containing 12 residual blocks \cite{He_2016_CVPR}:
\begin{equation}\label{eq:streak}
  \mathcal{L}_S = \mathcal{L}_{MSE}(\mathbf{ S}, \mathbf{S_{gt}}),
\end{equation}
where $\mathcal{L}_S$ represents the loss for learning rain streaks and $\mathbf{S}_{gt}$ is the groundtruths of a rain-streaks map.

\vspace{0.3cm}
\noindent \textbf{Learning Atmospheric Light}
The atmospheric light subnetwork learns to predict the global atmospheric light $ \mathbf{ A}$ only from the low-frequency component $\{\mathbf{I}_L\}^k$. This is because the low-frequency component does not contain rain streaks, where its specular reflection may significantly change the brightness of the input image and adversely affects the estimation of $ \mathbf{ A}$. This subnetwork is composed of 5 Conv+ReLU blocks appended with 2 fully-connected layers. The output vector $\mathbf{ A}$ is then upsampled to the size of the input image for the estimation of $ \mathbf{J}$ in Eq.~(\ref{eq:J_reconstruction}). The loss function for learning $\mathbf{A}$ is defined by:
\begin{equation}\label{eq:atm}
  \mathcal{L}_A = \mathcal{L}_{MSE}(\mathbf{ A}, \mathbf{A_{gt}}),
\end{equation}
where $\mathbf{A_{gt}}$ is the groundtruth of the atmospheric light.

\vspace{0.3cm}
\noindent \textbf{Learning Transmission}
We use an auto-encoder with skip connection to learn the transmission map $ \mathbf{T}$. We adopt the instance normalization \cite{Dmitry_2016_Arxiv} instead of batch normalization in the first two convolutional layers, as in our experiment, the latter performs poorly when the testing data has a significant domain gap from the training data. The loss function for learning $\mathbf{T}$ is defined as:
\begin{equation}\label{eq:trans}
  \mathcal{L}_T = \mathcal{L}_{MSE}(\mathbf{ T}, \mathbf{T_{gt}}),
\end{equation}
where $\mathbf{T_{gt}}$ refers to the ground-truth transmission map.

\vspace{0.3cm}
\noindent \textbf{Loss functions}
Based on the preceding, the overall loss function for the physics-based network to predict the physical parameters $\mathbf{\Theta}$ is:
\begin{equation}\label{eq:physics-based}
  \mathcal{L}_{\mathbf{\Theta}} = \lambda_{S} \mathcal{L}_{S} + \lambda_{A} \mathcal{L}_{A} + \lambda_{T} \mathcal{L}_{T},
\end{equation}
where $\lambda_{S}, \lambda_{A}$ and $\lambda_{T}$ are weighting factors for each loss. In our experiment, they are all set to $1$ since they are all MSE losses with the same scale.

\subsection{Stage 2: Model-Free Refinement}
The model-free refinement stage contains a conditional generative adversarial network. The generative network takes in the estimated image $\mathbf{J}$ and rain image $\mathbf{I}$ as input and produces the clean image $\mathbf{ C}$ to be assessed by the discriminative network. The overall loss function for the cGAN is:
\begin{eqnarray}
\min_G \max_D V(D,G) & = & \mathbb{E}_{\mathbf{C} \sim p(\mathbf{C},\mathbf{I})}[\log D(\mathbf{C|I})] \\
\nonumber  & + & \mathbb{E}_{\mathbf{J} \sim p_(\mathbf{J},\mathbf{I})}[\log (1 - D(G(\mathbf{J|I})))]
\label{eq:cGAN}
\end{eqnarray}
where $D$ represents the discriminative network and $G$ represents the generative network.

\vspace{0.3cm}
\noindent \textbf{Generative Network } The generative network is an autoencoder that contains 13 Conv-ReLU blocks, and skip connections are added to preserve more low-level image details. The goal of the generative network is to generate  a refined clean version $\mathbf{ C}$ that looks real and free from rain effect and artefacts produced by the previous stage. The input of this generator is $\mathbf{I}$ and $\mathbf{J}$. Since $\mathbf{J}$ is considerably sensitive to the estimation errors in the atmospheric light $\mathbf{ A}$, the generator may not be able to learn effectively. To improve the training, we inject the estimated atmospheric light $\mathbf{ A}$ into the generator as shown in Fig.~\ref{fig:arch}. In particular, we first embed $\mathbf{ A}$ into a higher dimensional space using two convolutions before concatenating the result with the encoder output of the generative network. This is done at the highest layer of the encoder where more global features are represented, because $\mathbf{ A}$ itself is a global property of the scene.

We also add MSE and perceptual losses \cite{Johnson2016Perceptual} for the training of the generative network. They are given by the first and second terms in the following loss function:
\begin{align*}
\mathcal{L}_C  = &\mathcal{L}_{MSE}(\mathbf{ C}, \mathbf{C_{gt}}) \\ \nonumber
& + \lambda_p \mathcal{L}_{MSE}(VGG(\mathbf{ C}), VGG(\mathbf{C_{gt}})) ,
\end{align*}
where $\lambda_p = 8$ in our experiment, and the perceptual loss is based on VGG16 pretrained on the ImageNet dataset.

Overall, the loss function for the generative network is:
\begin{align}
\mathcal{L}_{G} & =   \mathcal{L}_{C} +  \lambda_{GAN} \mathcal{L}_{GAN}(\mathbf{ C)},
\label{eq:total_loss}
\end{align}
where $\mathcal{L}_{GAN}(\mathbf{ C}) = log(1 - D(\mathbf{ C}))$ and the weighting parameter $\lambda_{GAN}$ is set to 0.01.

\vspace{0.3cm}
\noindent \textbf{Discriminative Network }  The discriminative network accepts the output of generative network and checks if it looks like a realistic clear scene. Since it is usually the distant scene that suffers loss of information, we want to make sure that the GAN focuses on these faraway parts of the scene. We first leverage the transmission map $\mathbf{ T}$ produced from the physics-based network and convert it to a relative depth map according to the relationship:
\begin{equation}\label{eq:trans_depth}
  \mathbf{T}(x) = \exp^{-\beta \mathbf{d}(x)},
\end{equation}
where $\mathbf{d}$ represents the scene depth and $\beta$ indicates the intensity of the veil or rain accumulation (in our experiment, $\beta$ is randomly sampled from a uniform distribution in [3, 4.2]). Then, we take the features from the 6\textsuperscript{th} Conv-ReLU layer of the discriminator and compute the MSE loss between these features and the depth map $ -\log (\mathbf{T})$ normalized to $[0, 1]$:
\begin{equation}\label{eq:Depth_loss}
  \mathcal{L}_{depth}(\mathbf{C}, \mathbf{T}) = \mathcal{L}_{MSE}( Conv(D(\mathbf{C})_{6}), Norm(-\log{\mathbf{T}})),
\end{equation}
where $D(\mathbf {C})_{m}$ represents the features at the $m$\textsuperscript{th} layer of the discriminator. We use the learnt depth map to weigh the features from the previous layer by multiplying them in an element-wise manner:
\begin{equation}
D(\mathbf {C})_{7} =   \mathbf{d} \odot D(\mathbf {C})_{6}.
\label{eq:disc_element-wise}
\end{equation}
Since faraway objects have higher depth values $\mathbf{d}$, the errors coming from these objects will be back-propagated to the generative network with greater weights during training.

The whole loss function of the discriminative network can be expressed as :
\begin{eqnarray}
\nonumber
\mathcal{L}_{D}  = &- log(D(\mathbf{C_{gt}})) - log(1 - D(\mathbf{ C})) \\
           + & \mathcal{L}_{depth}(\mathbf{C}, \mathbf{T}) + \mathcal{L}_{depth}(\mathbf{C}_{gt}, \mathbf{T}_{gt}).
\label{eq:total_generator}
\end{eqnarray}

\begin{algorithm}[t]
\caption{Algorithm for \textit{Outdoor-Rain} Rendering}
\label{alg:pipeline}
\begin{algorithmic}[1]
\State \textbf{Input: }Clean Image $\mathbf{C}$ and its depth map $\mathbf{D}$
\State $\mathbf{C}_{blur}(\mathbf{x}) = \mathsf{imgaussfilt}(\mathbf{C}(\mathbf{x}), \mathbf{\sigma_{C}}(\mathbf{x}))$. The smooth kernel varies according to depth: $\mathbf{\sigma_{C}}(\mathbf{x}) = 1.5  \mathbf{D}(\mathbf{x})$.
\State Generate 2D Noise map $\mathbf{N}$ with $\mu \sim - \textit{U}(0, 0.2) - 0.8 $, $\sigma \sim \textit{U}(0, 0.3) + 0.7 $
\State Rain Streaks map $\mathbf{S} = \mathsf{immotionfilt}(\mathbf{N}, l, \theta)$, parameter $l \sim \textit{U}(0, 40) + 20$, $\theta \sim \textit{U}(80, 100)$
\State Obtain Rain image $\mathbf{I}_S = \mathbf{S} + \mathbf{C}_{blur}$
\State Obtain Transmission $\mathbf{T} = \exp^{-\beta  \mathbf{D}}, \beta \sim \textit{U}(3, 4.2)$
\State Obtain $\mathbf{T}_{blur} = \mathsf{imgaussfilt}(\mathbf{T}, \sigma_{T})$, $\sigma_{T} \sim \mathcal{N} (5, 1.5)$.
\State Obtain global atmospheric light  $\mathbf{A} \sim \textit{U}(0.3, 0.8)$
\State \textbf{Output: } Rain Image $\mathbf{I} = \mathbf{T}_{blur} \mathbf{I}_R  + (1 - \mathbf{T}_{blur}) \mathbf{A}$
\end{algorithmic}
\end{algorithm}

\section{Implementation}

\subsection{Data Generation}
There are several large-scale synthetic datasets available for training deraining networks; however none of them contains rain accumulation effects. Hence, for the training of the physics-based stage, we create a new synthetic rain dataset named \textit{NYU-Rain}, using images from NYU-Depth-v2 \cite{Silberman:ECCV12} dataset as background. We render synthetic rain streaks and rain accumulation effects based on the provided depth information. These effects include the veiling effect caused by the water particles, as well as image blur (for details of the rain rendering process, see Algorithm 1). This dataset contains 16,200 image samples, out of which 13,500 images are used as the training set. For the training of the model-free refinement stage, we create another outdoor rain dataset on a set of outdoor clean images from \cite{Qian_2018_CVPR}, denoted as \textit{Outdoor-Rain}. In order to render proper rain streaks and rain accumulation effects as above, we estimate the depth of the scene using the state of the art single image depth estimation method \cite{monodepth17}. This dataset contains 9000 training samples and 1,500 validation samples.

\subsection{Training Details}
The proposed network is first trained in a stage-wise manner and then fine-tuned on an end-to-end basis. To train the physics-based stage on the \textit{NYU-Rain} dataset, we use Adam \cite{Kingma_2014_NIPS} optimizer with weight decay $10^{-4}$ and only supervise $\mathcal{L}_{\mathbf{\Theta}}$.  The learning rate is set to 0.001 initially and is divided by 2 after every 10 epochs until the 60\textsuperscript{th} epoch. To train the model-free refinement stage, we fix the parameters of the physics-based network and use the same optimizer and learning rate schedule as above. This model-free network is trained up to the 100\textsuperscript{th} epochs in this stage. Finally, we unfreeze the parameters in the physics-based network and fine-tune the entire model for a few thousand iterations. The entire network is implemented in Pytorch framework and will be made publicly available. \footnote{https://github.com/liruoteng/HeavyRainRemoval}

\section{Experimental Results}
In this section, we evaluate our algorithm with a few baseline methods on both the synthetic rain data and real rain data. For synthetic rain evaluation, we created a test datasets based on the test images from \cite{Qian_2018_CVPR} using the same rendering techniques in Algorithm 1, denoted as \textit{Test 1}. For a fair comparison with baselines, we combine the state of the art dehazing method \cite{NonLocalImageDehazing} with a series of state of the art rain streaks removal methods: (a) Deep detailed Network (DDN) \cite{Fu_2017_CVPR}, (b) DID-MDN method \cite{Zhang_2018_CVPR}, (c) RESCAN \cite{Li_2018_ECCV} method, and (d) JCAS \cite{Gu_2017_ICCV} method. In addition, we also compare with Pix2Pix GAN \cite{pix2pix2016} and CycleGAN \cite{CycleGAN2017} trained on the \textit{Outdoor-Rain} dataset.

\setlength{\tabcolsep}{2pt}
\begin{table}[]
\caption{A comparison on performance of estimated $\mathbf{S}$, $\mathbf{A}$, $\mathbf{T}$ and $\mathbf{J}$ among three different architectures on \textit{Test 1} data. }
\centering
\vspace{-10pt}
\resizebox{0.47\textwidth}{!}{%
\begin{tabular}{c|c|c|c|c|c}
\hline
\hline
Method   & Guidance Image                    & $\mathbf{J}$      & $\mathbf{S}$      & $\mathbf{T}$      & $\mathbf{A}$  \\
Metric             & {}                      & PSNR              & PSNR              & PSNR              & Error  \\
\hline
No Decomposition &  -                        & 10.87              & 23.65            & 14.95             & 0.212   \\		     		
\hline
Decomposition  & Input Image                 & 11.30              & 23.42     	     & 15.85         	 & 0.151  \\		
\hline
Decomposition & Residue Channel              & \textbf{13.83}    & \textbf{23.70}    & \textbf{19.48}    & \textbf{0.150}   \\		
\hline
\hline
\multicolumn{2}{c|}{Improvement over ``No Decomposition"}                 & 27.23 \%       & 0.21 \%        & 30.30 \%          & 29.25 \% \\
\hline
\hline
\end{tabular}}
\vspace{-5pt}
\label{table:decomp}
\end{table}

\begin{figure*}[t]
  \centering
  \subfloat[Input]{\includegraphics[width=0.13\linewidth]{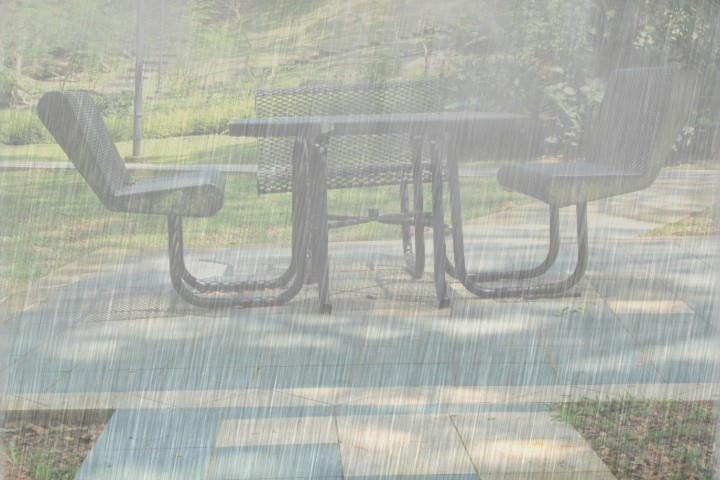}}
  \subfloat[DDN \cite{Fu_2017_CVPR} + \cite{NonLocalImageDehazing}]{\includegraphics[width=0.13\linewidth]{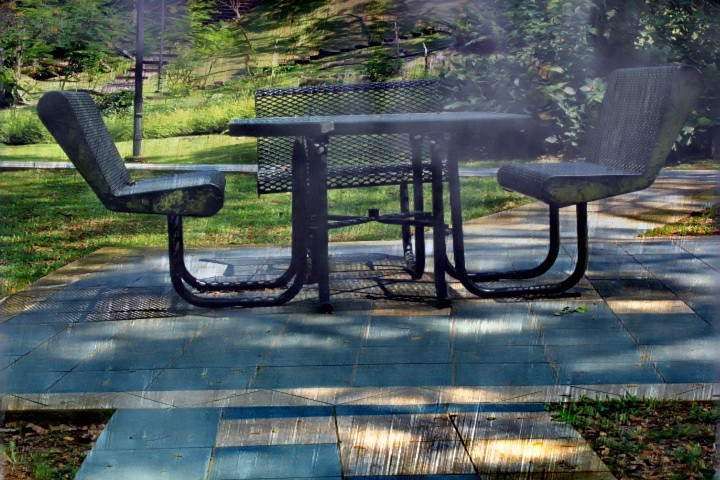}}
  \subfloat[DID \cite{Zhang_2018_CVPR} + \cite{NonLocalImageDehazing}]{\includegraphics[width=0.13\linewidth]{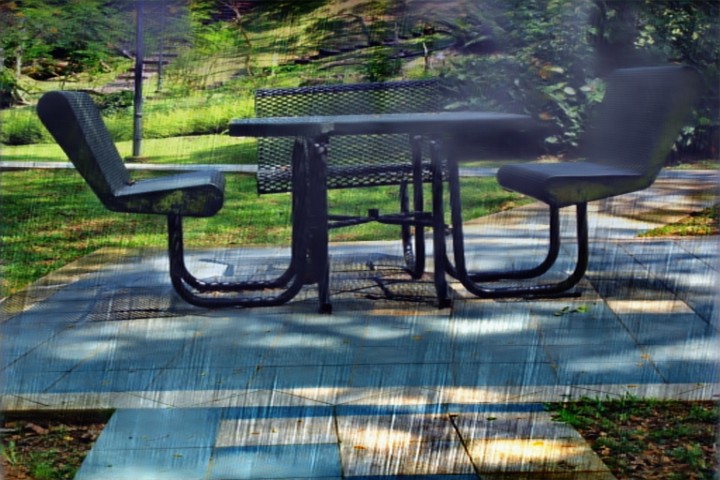}}
  \subfloat[RESCAN+ \cite{NonLocalImageDehazing}]{\includegraphics[width=0.13\linewidth]{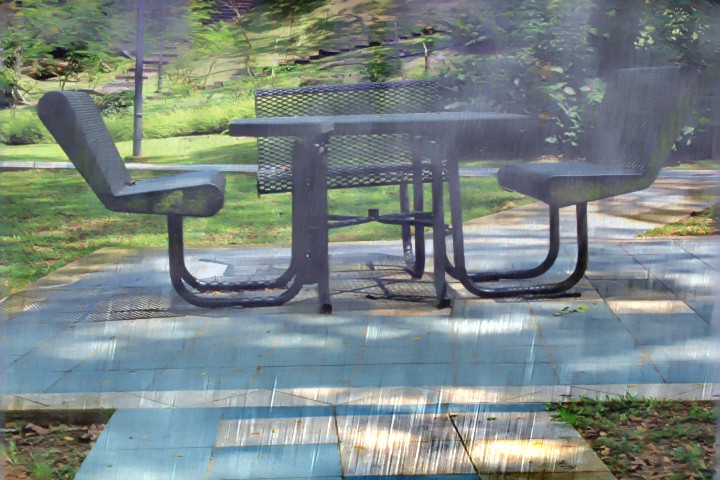}}
  \subfloat[Pix2Pix \cite{pix2pix2016}]{\includegraphics[width=0.13\linewidth]{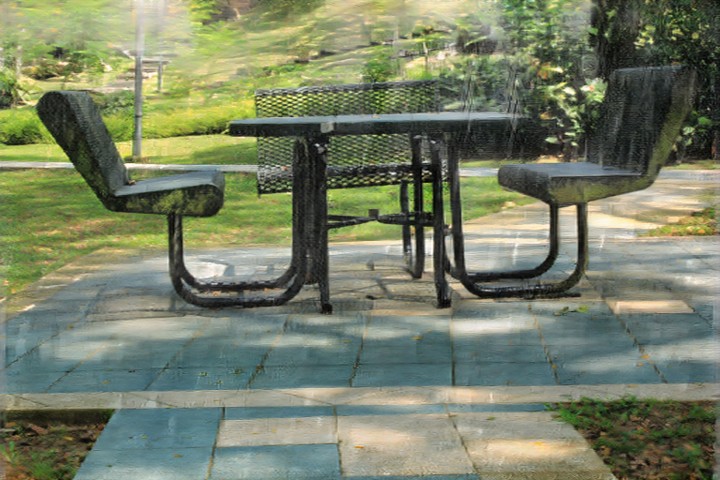}}
  \subfloat[CycleGAN \cite{CycleGAN2017}]{\includegraphics[width=0.13\linewidth]{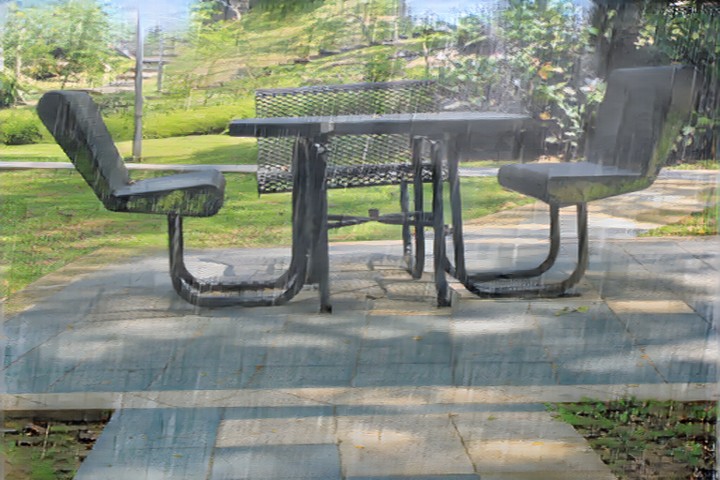}}
  \subfloat[Ours]{\includegraphics[width=0.13\linewidth]{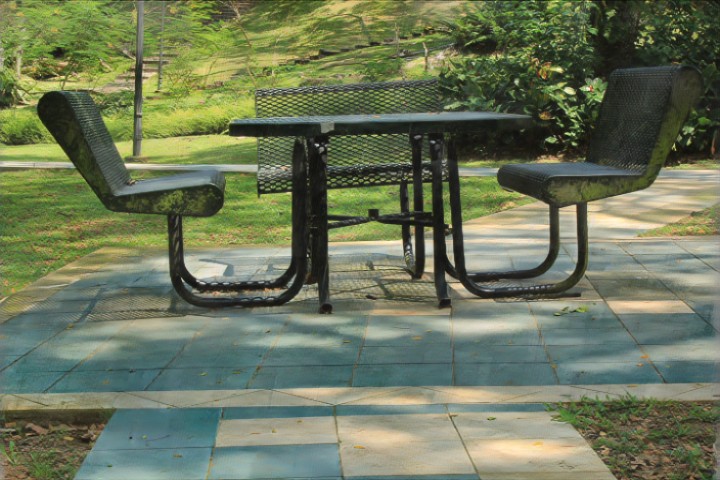}}
  \subfloat[Ground Truth]{\includegraphics[width=0.13\linewidth]{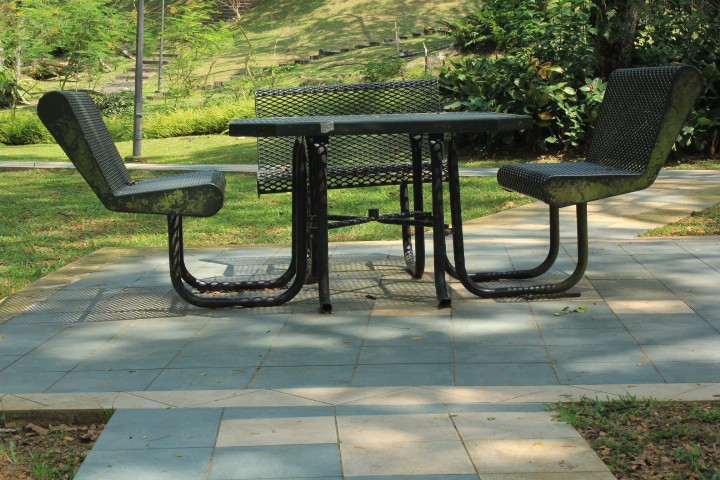}}
  \vspace{-5pt}
  \caption{A comparison of our algorithm with the baseline methods performed on \textit{Test 1} dataset. }\label{fig:Synthetic_Rain_Result}
  \vspace{-5pt}
\end{figure*}

\subsection{Ablation Study}

\noindent \textbf{Derain + Dehaze or Dehaze + Derain?}
The first ablation study evaluates the performance of combined dehazing and deraining methods in different order. We denote DeHaze First as \textit{DHF} and DeRain First as \textit{DRF}. We test these methods on \textit{Test 1}  dataset and  Table~\ref{table:quantitative} shows the quantitative results of these baseline methods in PSNR \cite{PSNR} and SSIM \cite{SSIM} metric. We will henceforth compare our method with the better pipeline.

\vspace{0.3cm}
\noindent \textbf{Decomposition Module}
To study the effectiveness of the decomposition module, we compared three different network architectures: (a) No decomposition module in the first stage, denoted as ``No Decomposition". (b). Decomposition module using input image as guidance image, denoted as ``Input-guided Decomposition". (c). We use the architecture proposed in this paper, named as ``Residue-guided Decomposition". We run these three methods on the testing dataset \textit{Test 1} and evaluate the estimated $\mathbf{S}$, $\mathbf{T}$ and the reconstructed image $\mathbf{J}$ in PSNR \cite{PSNR} metric. For atmospheric light $\mathbf{A}$, we evaluated the sum error against the ground-truth $\mathbf{A}_{gt}$: $Error = \sum_{i \in {r,g,b}} |{\mathbf{A}^{i} - \mathbf{A}_{gt}^{i}}|$. From the quantitative results shown in Table~\ref{table:decomp}, the decomposition operation significantly increases the accuracy of transmission estimation and thus improves the reconstructed image $\mathbf{J}$. Since
the decomposition guided by input image cannot fully separate rain streaks from the low-frequency component, the estimated $\mathbf{S}$ does not gain advantage. However, using the streak-free residue channel as guidance image, the transmission and atmospheric light will benefit from the streak-free low-frequency component, leading to further improvement on estimation.

\setlength{\tabcolsep}{2pt}
\begin{table}[]
\caption{A comparison of our algorithm with the baseline methods performed on \textit{Test 1} dataset. }
\centering
\vspace{-5pt}
\resizebox{0.33\textwidth}{!}{%
\begin{tabular}{c|c|cc}
\hline
\hline
\multicolumn{2}{c|}{Method}                                             & \multicolumn{2}{c}{Test 1}    \\
\hline
\multicolumn{2}{c|}{Metric}                                             & PSNR              & SSIM      \\
\hline
\multirow{2}{*}{JCAS \cite{Gu_2017_ICCV} + Dehaze}          & DHF       & 14.95         & 0.590         \\ %
                                                            & DRF       & 16.44         & 0.599        	\\ %
\hline
\multirow{2}{*}{DDN \cite{Fu_2017_CVPR} + Dehaze}           & DHF       & 13.36         & 0.583         \\ %
                                                            & DRF       & 15.68         & 0.640         \\
\hline
\multirow{2}{*}{DID-MDN \cite{Zhang_2018_CVPR} + Dehaze}    & DHF       & 14.17         & 0.577         \\ %
                                                            & DRF       & 12.58         & 0.471         \\ %
\hline
\multirow{2}{*}{RESCAN \cite{Li_2018_ECCV} + Dehaze}        & DHF       & 14.72         & 0.587         \\
                                                            & DRF       & 15.91         & 0.615         \\ %
\hline
\multicolumn{2}{c|}{Pix2Pix \cite{pix2pix2016}}                         & 19.09         & 0.710         \\
\hline
\multicolumn{2}{c|}{CycleGAN \cite{CycleGAN2017}}                       & 17.62         & 0.656         \\
\hline
\multicolumn{2}{c|}{No Decomposition + Stage 2}                         & 20.82         & 0.832         \\
\hline
\hline
\multicolumn{2}{c|}{Ours-$\mathbf{J}$}                                  & 20.05         & 0.779         \\
\hline
\multicolumn{2}{c|}{Ours-$\mathbf{C}$}                                  & \bf{21.56}    & \bf{0.855}  	 \\
\hline
\hline
\end{tabular}}
\vspace{-5pt}
\label{table:quantitative}
\end{table}

\begin{figure}[t]
  \centering
  \subfloat[Input]{\includegraphics[width=0.24\linewidth]{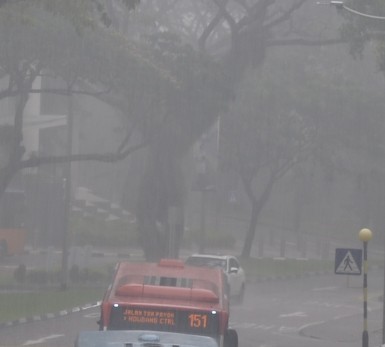}}
  \subfloat[$\mathbf{J}$]{\includegraphics[width=0.24\linewidth]{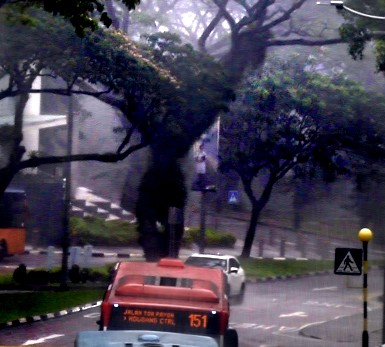}}
  \subfloat[$\mathbf{C}$]{\includegraphics[width=0.24\linewidth]{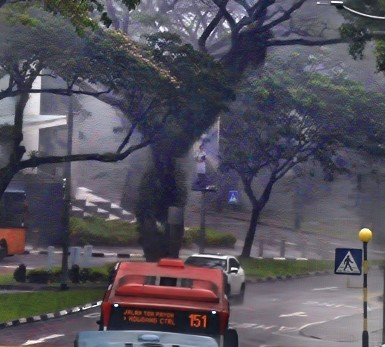}}
  \subfloat[$\mathbf{d}$]{\includegraphics[width=0.24\linewidth]{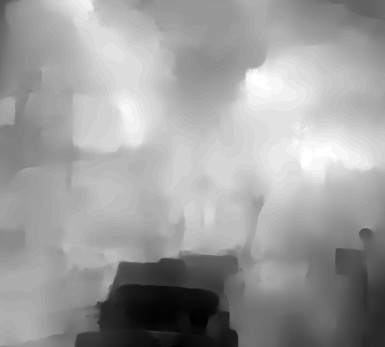}}
   \vspace{-5pt}
  \caption{The reconstructed image $\mathbf{J}$ produces darkened result on distant objects. The refinement network restores the details according to normalized depth map $\mathbf{d}$.}\label{fig:Analysis_J_C}
  \vspace{-5pt}
\end{figure}

\vspace{0.3cm}
\noindent \textbf{Study of Refinement Stage}
Fig.~\ref{fig:Analysis_J_C} shows the comparison between reconstructed image $\mathbf{J}$ and final output $\mathbf{C}$ produced by our network on real-world rain image. One can observe that there are dark regions around the distant tree are on image $\mathbf{J}$. The darkened result is one of the common problems in dehazing methods. Our refinement network is able to identify these areas and restore the contextual details of the distant tree with visually fine color according to the relative depth map $\mathbf{d}$ converted from estimated transmission map $\mathbf{T}$ using Eq.~(\ref{eq:trans_depth}).

\subsection{Synthetic Rain Analysis}
Table~\ref{table:quantitative} demonstrates the quantitative performance of our algorithm compared with the baseline methods in PSNR \cite{PSNR} and SSIM \cite{SSIM} metrics. Fig.~\ref{fig:Synthetic_Rain_Result} shows the qualitative results produced by our algorithm and other baseline methods. Here, we choose the better performed result between dehaze+derain and derain+dehaze for those rain streaks removal methods. \cite{Gu_2017_ICCV}\cite{Fu_2017_CVPR}\cite{Li_2018_ECCV}\cite{Zhang_2018_CVPR}. Note that directly using GAN method such as \cite{pix2pix2016} \cite{CycleGAN2017} does not produce appropriate solution for this image enhancement problem since these generative models can sometimes generate fake results as shown in the first example (top part) of Fig.\ref{fig:Synthetic_Rain_Result}.

\begin{figure*}[!ht]
  \centering
  \captionsetup{position=top}
  \vspace{-5pt}
  \subfloat[Input]{\includegraphics[width=0.16\linewidth]{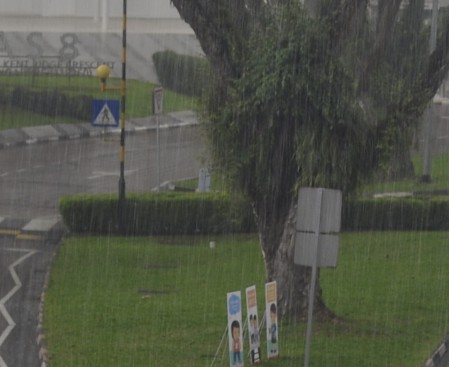}}
  \subfloat[Ours]{\includegraphics[width=0.16\linewidth]{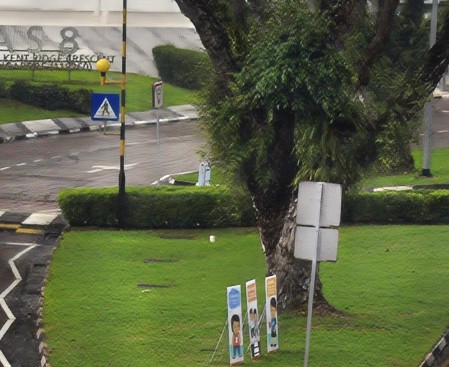}}\vspace{0.5pt}
  \subfloat[CycleGAN \cite{CycleGAN2017}]{\includegraphics[width=0.16\linewidth]{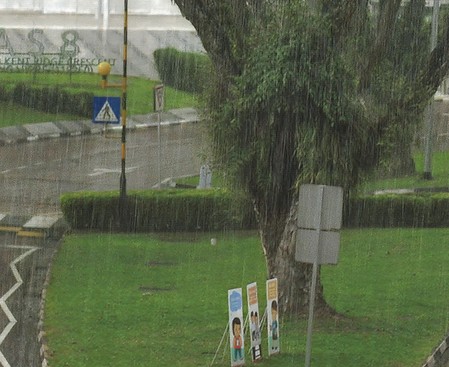}}
  \subfloat[\cite{NonLocalImageDehazing}+DID-MDN \cite{Zhang_2018_CVPR}]{\includegraphics[width=0.16\linewidth]{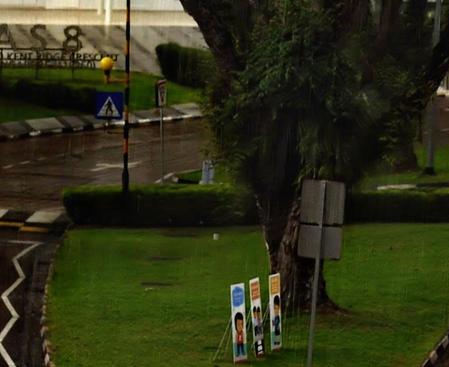}}
  \subfloat[RESCAN \cite{Li_2018_ECCV} + \cite{NonLocalImageDehazing}]{\includegraphics[width=0.16\linewidth]{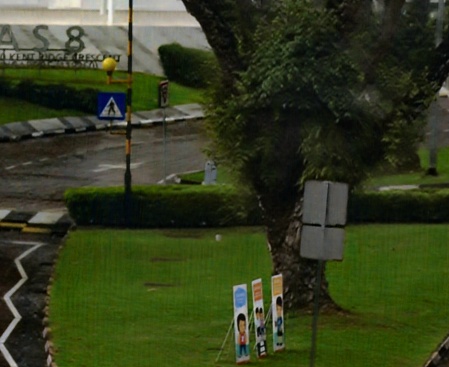}}
  \subfloat[Reference]{\includegraphics[width=0.16\linewidth]{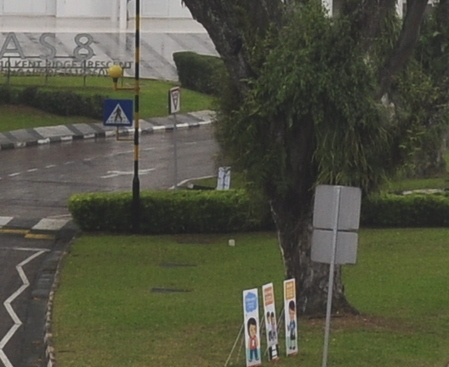}} \\
  \vspace{-12.6pt}
  \subfloat{\includegraphics[width=0.16\linewidth]{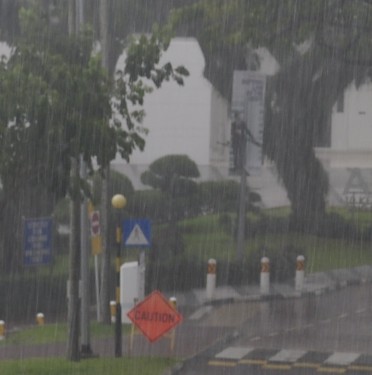}}
  \subfloat{\includegraphics[width=0.16\linewidth]{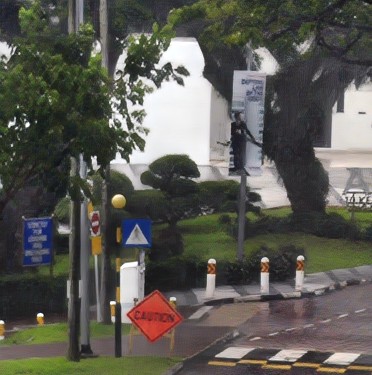}} \vspace{0.5pt}
  \subfloat{\includegraphics[width=0.16\linewidth]{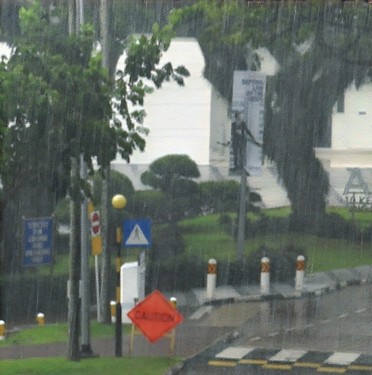}}
  \subfloat{\includegraphics[width=0.16\linewidth]{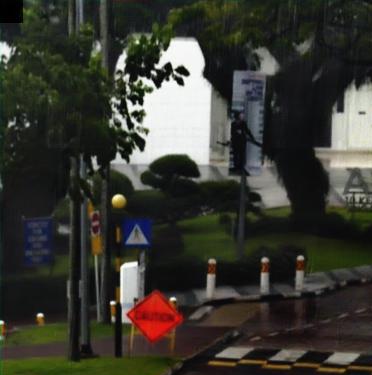}}
  \subfloat{\includegraphics[width=0.16\linewidth]{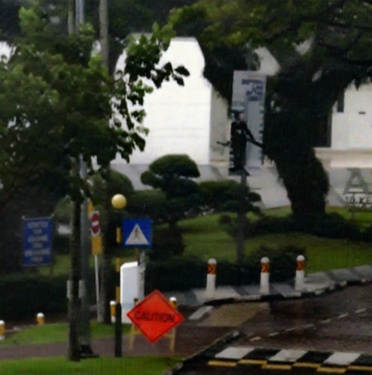}}
  \subfloat{\includegraphics[width=0.16\linewidth]{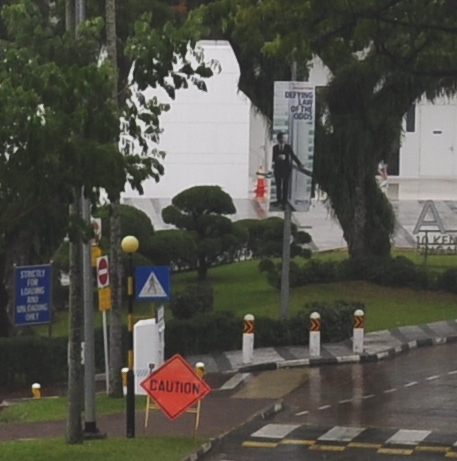}}   \\
  \vspace{-12.7pt}
  \subfloat{\includegraphics[width=0.16\linewidth]{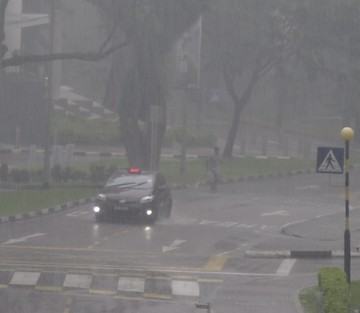}}
  \subfloat{\includegraphics[width=0.16\linewidth]{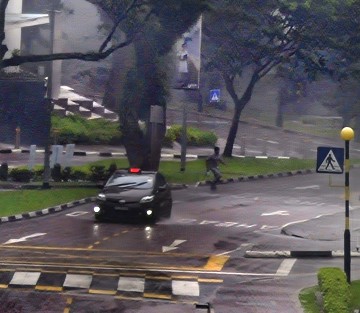}}\vspace{0.5pt}
  \subfloat{\includegraphics[width=0.16\linewidth]{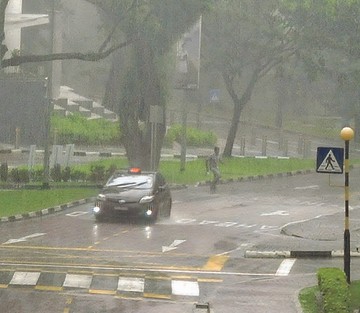}}
  \subfloat{\includegraphics[width=0.16\linewidth]{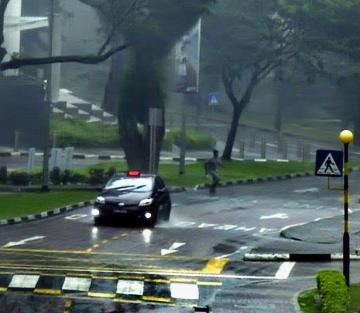}}
  \subfloat{\includegraphics[width=0.16\linewidth]{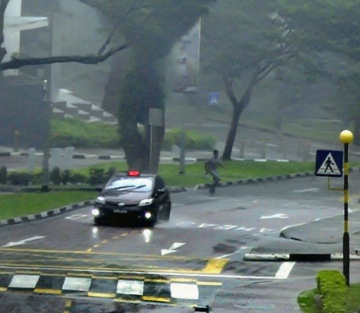}}
  \subfloat{\includegraphics[width=0.16\linewidth]{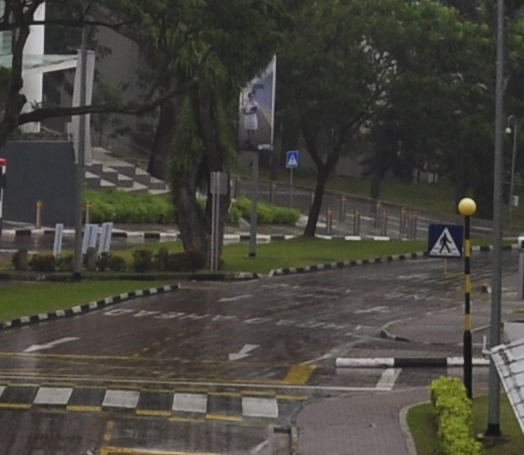}}    \\
  \addtocounter{subfigure}{-24}
  \vspace{-12.6pt}
  \subfloat{\includegraphics[width=0.16\linewidth]{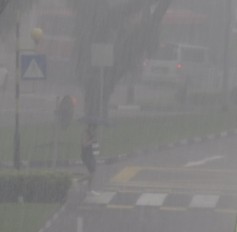}}
  \subfloat{\includegraphics[width=0.16\linewidth]{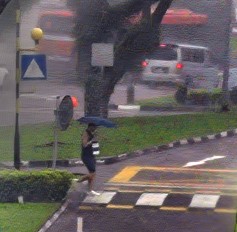}} \vspace{0.5pt}
  \subfloat{\includegraphics[width=0.16\linewidth]{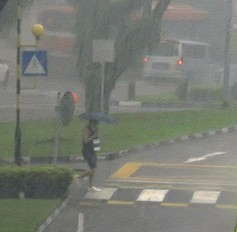}}
  \subfloat{\includegraphics[width=0.16\linewidth]{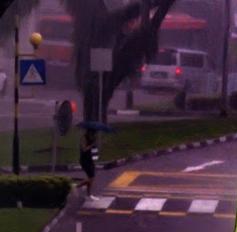}}
  \subfloat{\includegraphics[width=0.16\linewidth]{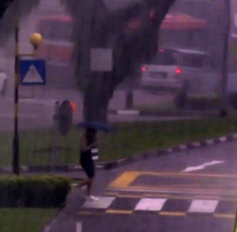}}
  \subfloat{\includegraphics[width=0.16\linewidth]{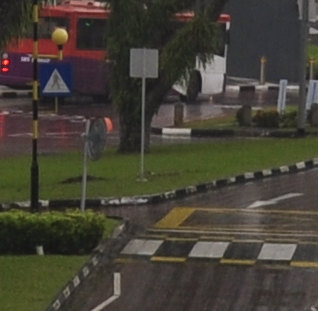}}\\
  \vspace{-12.6pt}
  \subfloat{\includegraphics[width=0.16\linewidth]{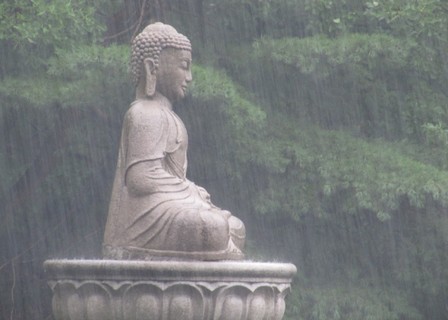}}
  \subfloat{\includegraphics[width=0.16\linewidth]{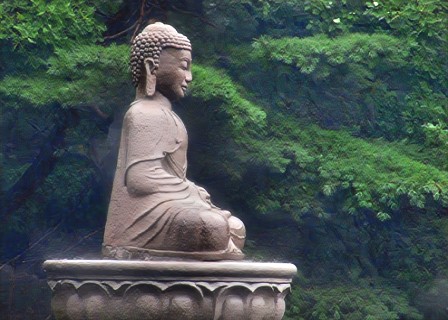}} \vspace{0.1pt}
  \subfloat{\includegraphics[width=0.16\linewidth]{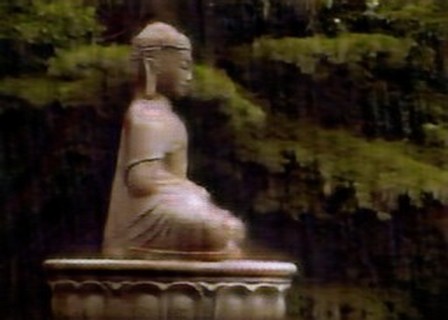}}
  \subfloat{\includegraphics[width=0.16\linewidth]{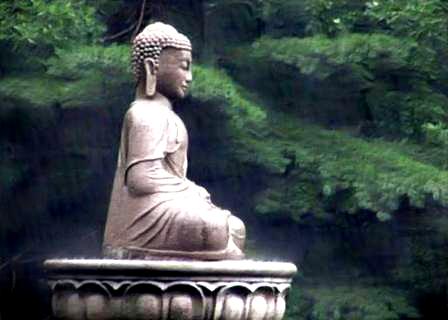}}
  \subfloat{\includegraphics[width=0.16\linewidth]{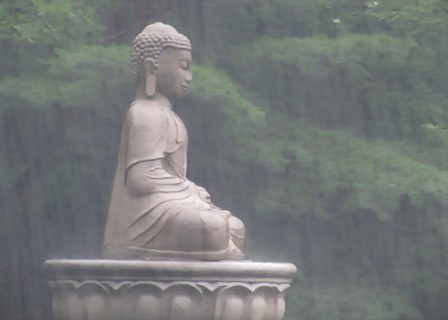}}
  \subfloat{\includegraphics[width=0.16\linewidth]{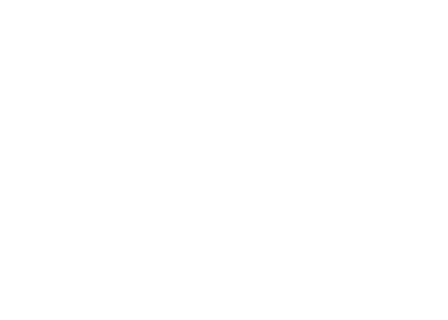}}  \\
  \vspace{-5pt}
  \caption{A comparison of our algorithm with baseline methods on real-world rain scenes. The reference images are other pictures taken just after rains. From top to bottom, the rain becomes more and more severe. (Zoom-in to view details). }\label{fig:Real_Rain_Result}
  \vspace{-25pt}
\end{figure*}

\begin{figure*}
  \centering
  \includegraphics[width=0.30\linewidth]{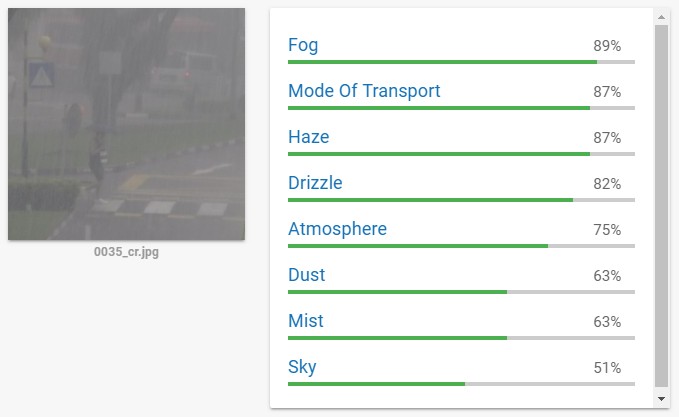}
  \includegraphics[width=0.30\linewidth]{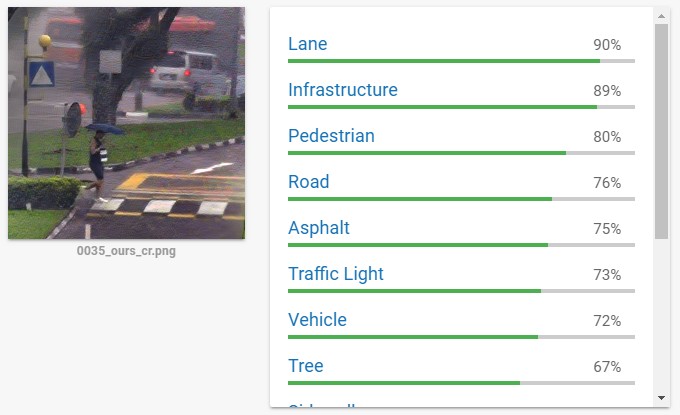}
  \raisebox{-0.1\height}{\includegraphics[width=0.35\linewidth]{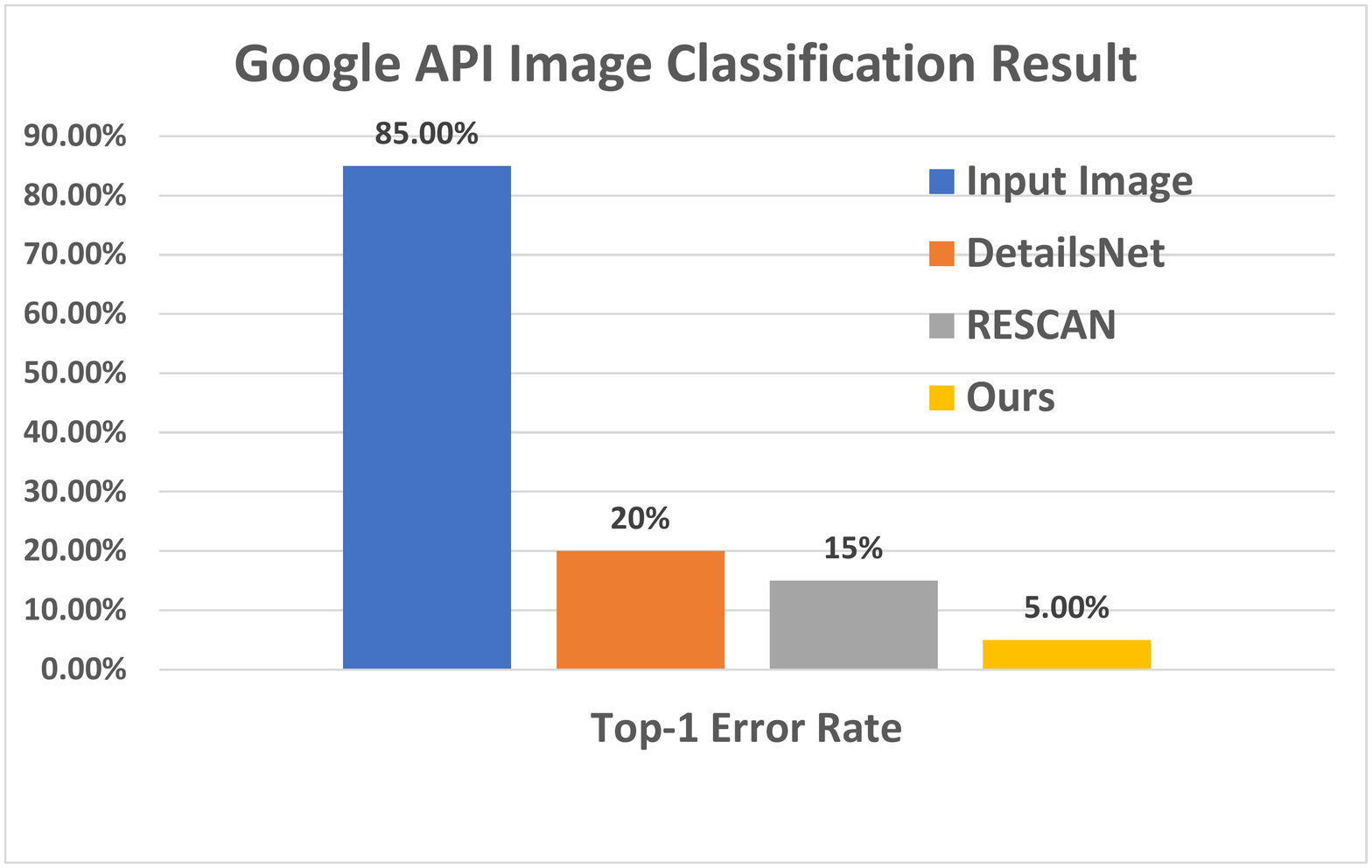}}
  \vspace{-20pt}
  \caption{ Object recognition results for the input rain image and our results respectively. We test 20 sets of rain and derain images of ours and baseline methods \cite{Li_2018_ECCV,Fu_2017_CVPR}. We record the top-1 error rate on the right bar chart. }\label{fig:Application}
  \vspace{-10pt}
\end{figure*}

\subsection{Real-world Rain Analysis}
\noindent \textbf{Qualitative Result}
Fig.~\ref{fig:Real_Rain_Result} shows the qualitative comparison between our method and other baseline methods. For the baseline methods under moderate rain scenes, the haze removal component usually produces dark results and the rain removal components inevitably damage the background details, resulting in blurred image. (e.g. the tree leaves and the lamp poles in Fig.~\ref{fig:Real_Rain_Result} Row 1,2). In the case of heavy rain, these baseline methods fail to remove the rain streaks effectively due to the presence of strong rain accumulation (Fig.~\ref{fig:Real_Rain_Result} Row 5). In addition, the state of the art haze removal method cannot effectively remove the veiling effect. One can still observe hazy effect at the remote area of the baseline results (row 4 of Fig.~\ref{fig:Real_Rain_Result}). Thanks to the depth guided GAN, our method is able to identify the remote areas and remove the proper amount of veiling effect.

\vspace{0.3cm}
\noindent \textbf{Application}
In order to provide the evidence that our image restoration method will benefit outdoor computer vision applications, we employ Google Vision API object recognition system to evaluate our results. Fig.~\ref{fig:Application} shows the screenshots of the results produced by Google API. We test 20 sets of real rain images and derained images of our method and baseline methods \cite{Fu_2017_CVPR,Li_2018_ECCV}. We report the classification results of top-1 error rate. As one can see, our method significantly improve the recognition results and outperforms other baseline methods.

\section{Conclusion}
We propose a novel 2-stage CNN that is able to remove rain streaks and rain accumulation simultaneously. In the first physics-based stage, a new streak-aware decomposition module is introduced to decompose the entangled rain streaks and rain accumulation for better joint feature extraction. Scene transmission and atmospheric light are also estimated to provide necessary depth and light information for second stage. We propose a conditional GAN in the refinement stage that takes in the reconstructed image from previous level and produce the final clean images. Comprehensive experimental evaluations show that our method outperforms the baselines on both synthetic and real rain data. 

\clearpage

{\small
\bibliographystyle{ieee}
\bibliography{egbib}
}

\end{document}